\documentclass{article}
\usepackage{amsthm}

\usepackage[preprint]{neurips_2026}
\newtheorem{theorem}{Theorem}
\usepackage{amssymb}
\usepackage{tabularx}
\usepackage{listings}
\lstdefinestyle{asmstyle}{
    basicstyle=\ttfamily\scriptsize,
    numbers=left,
    numberstyle=\tiny\color{gray},
    numbersep=5pt,
    frame=none,
    breaklines=true,
    tabsize=4,
    showstringspaces=false,
    xleftmargin=10pt,
    commentstyle=\color{commentgray},
    escapeinside={(*@}{@*)},
    columns=flexible,
    keepspaces=true,
}

\usepackage{amsmath}
\usepackage{enumitem}
\usepackage{pifont}
\usepackage{graphicx}

\newtheorem{proposition}{Proposition}

\usepackage[table]{xcolor} 
\usepackage{algorithm}
\usepackage{algpseudocode}
\usepackage{wrapfig}

\definecolor{darkgreen}{rgb}{0.0, 0.5, 0.0} 
\definecolor{verylightgray}{rgb}{0.97, 0.97, 0.97}
\definecolor{darkorange}{rgb}{0.8, 0.4, 0.0}

\newcommand{\cmark}{\textcolor{darkgreen}{\scalebox{1}[1.0]{\ding{51}}}}
\newcommand{\xmark}{\textcolor{red}{\ding{55}}}  

\newcommand{\res}[2]{$#1{\color{gray}\scriptstyle\,\pm#2}$}
\newcommand{\best}[2]{$\mathbf{#1}{\color{gray}\scriptstyle\,\pm#2}$}

\usepackage[utf8]{inputenc} 
\usepackage[T1]{fontenc}    
\usepackage{hyperref}       
\usepackage{url}            
\usepackage{booktabs}       
\usepackage{amsfonts}       
\usepackage{nicefrac}       
\usepackage{microtype}      

\title{CEPO: RLVR Self-Distillation using \\ \underline{C}ontrastive \underline{E}vidence \underline{P}olicy \underline{O}ptimization}

\author{%
  Ahmed Heakl\textsuperscript{1}\quad
  Abdelrahman M.\ Shaker\textsuperscript{1} \quad
  Youssef Mohamed\textsuperscript{1} \quad
  Rania Elbadry\textsuperscript{1} \\[2pt]
  \textbf{Omar Fetouh\textsuperscript{1} \quad
  Fahad Shahbaz Khan\textsuperscript{1,2} \quad
  Salman Khan\textsuperscript{1,3}} \\[6pt]
  \textsuperscript{1}MBZUAI \quad
  \textsuperscript{2}Linköping University \quad
  \textsuperscript{3}Australian National University 
}

\begin{document}

\maketitle

\begin{abstract}
When a model produces a correct solution under reinforcement learning with verifiable rewards (RLVR), every token receives the same reward signal regardless of whether it was a decisive reasoning step or a grammatical filler. A natural fix is to condition the model on the correct answer as a teacher, identifying tokens it would have generated differently had it known the answer. Prior work shows this either corrupts training by leaking the answer into the gradient, or produces a weak signal that cannot distinguish decisive steps from filler, since both look equally surprising relative to the model's baseline.
We propose Contrastive Evidence Policy Optimization (CEPO), which asks a sharper question at every token: not just ``does the correct answer favor this token?'' but ``does the correct answer favor it \emph{while} the wrong answer disfavors it?'' A token satisfying both is a genuine reasoning step; one satisfying neither is filler. The wrong-answer teacher is constructed from rejected rollouts already in the training batch, incurring no additional sampling cost. We prove CEPO inherits all structural safety guarantees of the prior state of the art while strictly sharpening credit at decisive tokens, with the improvement vanishing exactly at filler positions. Empirically, CEPO achieves 43.43\% and 60.56\% average accuracy across five multimodal mathematical reasoning benchmarks at 2B and 4B scale, respectively, versus 41.17\% and 57.43\% for GRPO under identical training budgets. Distribution-matching self-distillation methods (OPSD, SDPO) fall below the untrained baseline, empirically confirming the information leakage our theory predicts. Our code is available at \url{https://github.com/ahmedheakl/CEPO}.
\end{abstract}

\section{Introduction}
\label{sec:intro}

\begin{wrapfigure}{r}{0.50\linewidth}
    \vspace{-1.2em}
    \centering
    \includegraphics[width=\linewidth]{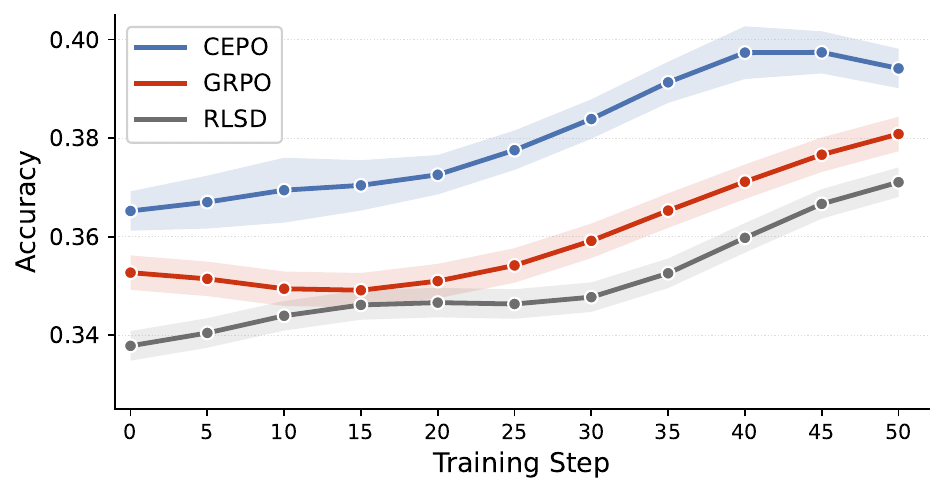}
    \caption{\textbf{Accuracy over 50 training steps.} CEPO improves faster than GRPO and RLSD, reaching its largest gap around step 40 before partially converging by the final checkpoint.}
    \label{fig:training}
    \vspace{-1.0em}
    
\end{wrapfigure}

Reinforcement learning with verifiable rewards (RLVR) has become the dominant paradigm for post-training large language models to reason~\citep{shao2024deepseekmath, guo2025deepseekr1, yang2025qwen3}. The core loop is simple: sample rollouts from the current policy, score them against a verifier, and update the policy to increase the probability of correct trajectories. Group Relative Policy Optimization~(GRPO)~\citep{shao2024deepseekmath} operationalizes this at scale by eliminating the value network entirely, normalizing rewards within groups of sampled responses to obtain sequence-level advantages. Yet the simplicity that makes GRPO practical also makes it blunt: every token in a correct trajectory receives the same positive advantage, and every token in a wrong one receives the same negative signal. The credit assignment problem, \emph{which tokens actually mattered?}, is left entirely unresolved.

This is not a minor inefficiency. In mathematical reasoning, a single arithmetic error or a single correct inferential step can determine the outcome of an entire chain-of-thought~\citep{kazemnejad2024vineppo, guo2025spo}. Uniform credit assignment wastes gradient signal on filler tokens (connectives, formatting, boilerplate) while underweighting the few decisive tokens that distinguish correct from incorrect reasoning. The result is slow convergence, noisy updates, and poor sample efficiency, problems that worsen as reasoning chains grow longer and sparser in decision-relevant content~\citep{zhang2026survey}. Figure~\ref{fig:training} illustrates this empirically, with CEPO improving faster than GRPO and RLSD early in training.

A natural fix is to condition the model on the correct answer $r^+$ as its own teacher, using the resulting distribution
$P_T^+(\cdot \mid x, r^+)$ as a dense, token-level training signal. On-policy self-distillation methods~\citep{zhao2026opsd,
hubotter2026sdpo, penaloza2026pidistill} pursue exactly this, minimizing a per-token divergence between $P_T^+$ and the student $P_S$ over on-policy rollouts. \citep{yang2026rlsd} showed this is structurally unsafe: the gradient of any divergence objective decomposes into a benign component and a harmful deviation with variance proportional to $I(Y_t; R^+ \mid X)$. As training progresses the benign signal vanishes and the deviation dominates, driving the model to encode spurious $x\!\to\!r^+$ correlations, a pathology termed \emph{information leakage} that is irreducible regardless of implementation details. 

RLSD~\citep{yang2026rlsd} resolved leakage by evaluating the evidence ratio $P_T^+(y_t)/P_S(y_t)$ only at the sampled token, under a stop-gradient, using it solely to modulate the \emph{magnitude} of the GRPO advantage while keeping its sign anchored to the verifier. No vocabulary-wide sum over $r$-conditioned weights appears in the gradient, so privileged information cannot redirect gradient flow. This is a sound structural recipe for safe self-distillation, but structural safety is not the same as signal quality. We identify three specific limitations of RLSD's evidence ratio. The denominator $P_S(y_t)$ reflects base-rate fluency, not semantic relevance, so a common token suppresses the ratio regardless of how strongly $r^+$ favors it (\textbf{fluency confound}). For wrong trajectories, the signal penalizes tokens that $r^+$ would have supported, indirect, with no explicit grounding in what $r^-$ predicts (\textbf{asymmetric negative}). Most critically, $P_T^+/P_S$ cannot distinguish a filler token that both the correct and wrong answers support equally from a decisive reasoning step that $r^+$ supports while $r^-$ actively disfavors; both receive identical weight (\textbf{one-sided evidence}).

We propose \textbf{Contrastive Evidence Policy Optimization (CEPO)}, which replaces $P_T^+/P_S$ with the contrastive ratio $P_T^+/P_T^-(y_t)$, where $P_T^-$ is the model conditioned on a wrong answer drawn from rejected rollouts already in the training batch. The student prior $P_S$ cancels entirely, eliminating the fluency confound by construction. The contrastive ratio admits a clean Bayesian interpretation as the \emph{differential belief update}: how much token $y_t$ simultaneously raises posterior belief in $r^+$ and lowers it for $r^-$. Decisive reasoning steps score high; filler tokens score near unity.

We prove CEPO preserves all structural safety guarantees of RLSD: direction anchoring ($\operatorname{sign}(\hat{A}_t) = \operatorname{sign}(A)$ for all tokens) and leakage-free gradients (no vocabulary-wide $r$-conditioned sum). When $P_T^-(y_t) = P_S(y_t)$, CEPO reduces exactly to RLSD, making RLSD a limiting case when the wrong-answer teacher carries no information. Beyond these guarantees, Proposition~\ref{prop:sharpness} gives exact necessary and sufficient conditions for CEPO to assign strictly sharper credit than RLSD at any token: for correct trajectories, sharpness holds precisely when $P_T^-(y_t) < P_S(y_t)$, a condition we validate empirically concentrates at arithmetically and inferentially decisive positions rather than at filler.

\paragraph{Contributions.}
\begin{enumerate}
\item We identify three concrete limitations of RLSD's evidence ratio: the fluency confound, asymmetric negative signal, and one-sided evidence.
\item We propose CEPO, replacing $P_T^+/P_S$ with $P_T^+/P_T^-$, with a Bayesian interpretation as the differential belief update which inherits all structural safety guarantees of RLSD while strictly generalizing it.
\item We derive exact conditions under which CEPO sharpens credit relative to RLSD and validate empirically that these concentrate at semantically decisive token positions.
\item We demonstrate accuracy improvements of 3.7\% and 2.2\% over base at 2B and 4B scale across five multimodal mathematical reasoning benchmarks.
\end{enumerate}

\section{Related Work}
\label{sec:related}

\begin{wraptable}{r}{0.52\linewidth}
\centering
\vspace{-1.7em}
\caption{\textbf{Comparison of credit assignment methods.}
  \textit{Priv.}: uses privileged info at training time.
  \textit{Leak-free}: no vocabulary-wide $r$-conditioned gradient.
  \textit{Contr.}: uses both positive and negative references.
  \textit{No Aux.}: requires no auxiliary network.}
\label{tab:related}
\scriptsize
\rowcolors{2}{verylightgray}{white}
\resizebox{\linewidth}{!}{
\begin{tabular}{lcccc}
\toprule
\textbf{Method} & \textbf{Priv.} & \shortstack{\textbf{Leak-}\\\textbf{free}} & \textbf{Contr.} & \shortstack{\textbf{No}\\\textbf{Aux.}} \\
\midrule
GRPO~\citep{shao2024deepseekmath}   & \xmark & --- & \xmark & \cmark \\
PPO~\citep{schulman2017ppo}         & \xmark & --- & \xmark & \xmark \\
VinePPO~\citep{kazemnejad2024vineppo} & \xmark & --- & \xmark & \cmark \\
SPO~\citep{guo2025spo}              & \xmark & --- & \xmark & \cmark \\
PRM~\citep{lightman2023lets}        & \xmark & --- & \xmark & \xmark \\
DPO/cDPO~\citep{rafailov2023dpo,lin2024cdpo} & --- & --- & \cmark & \cmark \\
\midrule
OPSD~\citep{zhao2026opsd}           & \cmark & \xmark & \xmark & \cmark \\
SDPO~\citep{hubotter2026sdpo}       & \cmark & \xmark & \xmark & \cmark \\
HDPO~\citep{ding2026hdpo}           & \cmark & \xmark & \xmark & \cmark \\
RLSD~\citep{yang2026rlsd}           & \cmark & \cmark & \xmark & \cmark \\
\midrule
\rowcolor{white}
\textbf{CEPO (ours)} & \cmark & \cmark & \cmark & \cmark \\
\bottomrule
\end{tabular}
}
\vspace{-1.0em}
\end{wraptable} 

\paragraph{RLVR and the credit assignment bottleneck.}
Reinforcement learning with verifiable rewards trains language models 
by scoring sampled rollouts against a deterministic 
verifier~\citep{guo2025deepseekr1}. GRPO~\citep{shao2024deepseekmath} 
eliminates the value network by normalizing rewards within a rollout 
group, and extensions such as DAPO~\citep{yu2025dapo} improve 
exploration stability. All methods in this family assign uniform 
sequence-level advantages: every token in a correct trajectory receives 
the same signal regardless of its contribution. Token-level methods 
address this gap either through Monte Carlo re-simulation, as in 
VinePPO~\citep{kazemnejad2024vineppo} and SPO~\citep{guo2025spo}, or 
through a separately trained process reward model 
(PRM;~\citep{lightman2023lets,setlur2024rewarding}). Both families 
appear in the top block of Table~\ref{tab:related}: they improve credit 
assignment without privileged information but either require expensive 
re-simulation or an auxiliary network.

\paragraph{On-policy self-distillation with privileged information.}
A natural alternative is to condition the model on the correct answer 
$r^+$ as its own teacher, producing a dense token-level signal at no 
auxiliary network cost. OPSD~\citep{zhao2026opsd} minimizes the 
per-token KL divergence between the privileged teacher $P_T^+$ and 
the student; SDPO~\citep{hubotter2026sdpo} extends this with 
Jensen-Shannon divergence and EMA teacher stabilization; and 
HDPO~\citep{ding2026hdpo} applies the same recipe specifically to 
prompts where all rollouts fail. As shown by~\citep{yang2026rlsd}, 
any method that uses $P_T^+$ as a distributional target produces 
gradients containing a vocabulary-wide sum of $r$-conditioned weights, 
a structural source of information leakage whose variance is 
irreducible regardless of implementation. These methods are marked 
\textit{Priv.} but not \textit{Leak-free} in Table~\ref{tab:related}, 
and we confirm their degradation empirically in \S\ref{sec:results}. 
The closest work to the contrastive direction within the DPO 
family~\citep{rafailov2023dpo} is cDPO~\citep{lin2024cdpo}, which 
identifies critical tokens via contrastive estimation, but it operates 
offline on fixed response pairs under a sequence-level implicit reward 
rather than within the RLVR loop.

RLSD~\citep{yang2026rlsd} resolves leakage by evaluating the teacher 
signal only at the sampled token under a stop-gradient, using the 
evidence ratio $P_T^+(y_t)/P_S(y_t)$ solely to modulate the 
\textit{magnitude} of the GRPO advantage while anchoring its direction 
to the verifier. This makes RLSD both \textit{Priv.} and 
\textit{Leak-free}, which no prior method achieves. However, the 
denominator $P_S(y_t)$ conflates reasoning importance with base-rate 
fluency, the negative signal for wrong trajectories is indirect, and 
the ratio cannot distinguish a decisive reasoning step from filler when 
both have the same $P_T^+/P_S$ value.

\section{Method}
\label{sec:method}

\begin{figure}[t]
    \centering
    \includegraphics[width=\linewidth]{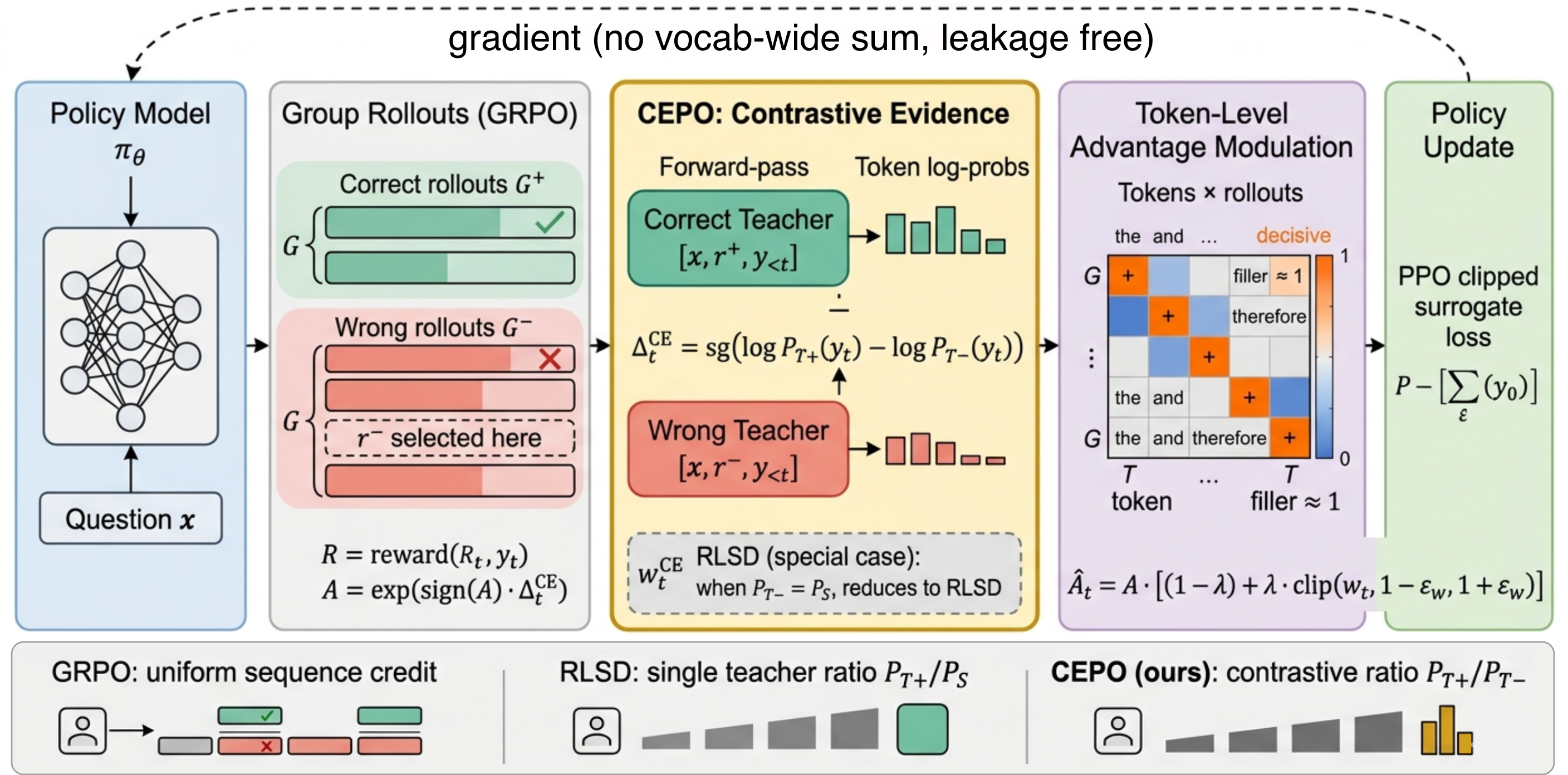}
    \caption{\textbf{CEPO training pipeline and its relationship to GRPO and RLSD.}
    Given a question $x$, the policy $\pi_\theta$ produces $G$ rollouts that are
    partitioned into correct ($G^+$) and wrong ($G^-$) sets by a verifiable reward.
    CEPO conditions two frozen teachers on a sampled correct rationale
    $r^+ \in G^+$ and rejected rationale $r^- \in G^-$, and defines a per-token
    \emph{contrastive evidence delta}
    $\Delta_t^{\mathrm{CE}}$ that amplifies advantage at decisive tokens (large $|\Delta_t^{\mathrm{CE}}|$)
    while leaving filler tokens near unit weight. Then, the token-level modulated advantage is plugged into a standard PPO-clipped surrogate.}
    \label{fig:main}
\end{figure}

\subsection{Preliminaries}
\label{sec:prelim}

Let $\pi_\theta$ be an autoregressive language model with parameters $\theta$ and vocabulary $\mathcal{V}$, trained on $\mathcal{S} = \{(x_i, r_i^+)\}_{i=1}^N$ where $r_i^+$ is a verifiable correct answer. A deterministic verifier $R: \mathcal{X} \times \mathcal{Y} \to \{0,1\}$ scores responses. GRPO~\citep{shao2024deepseekmath} samples $G$ rollouts per question and computes a normalized sequence-level advantage:
\begin{equation}
\label{eq:grpo_advantage}
A^{(i)} = \frac{R(x, y^{(i)}) - \mu_G}{\sigma_G},
\end{equation}
partitioning rollouts into correct ($\mathcal{G}^+$) and wrong ($\mathcal{G}^-$) subsets. We define three next-token distributions sharing parameters $\theta$ but differing in context:
\begin{align}
P_S(y_t) &\triangleq \pi_\theta(y_t \mid x, y_{<t}), \quad
P_T^+(y_t) \triangleq \pi_\theta(y_t \mid x, r^+, y_{<t}), \quad
P_T^-(y_t) \triangleq \pi_\theta(y_t \mid x, r^-, y_{<t}),
\end{align}
denoting the student, correct teacher, and wrong teacher respectively. We write $\operatorname{sg}(\cdot)$ for the stop-gradient operator.

\subsection{Background: Leakage in Self-Distillation and the RLSD Fix}
\label{sec:background}

Methods such as OPSD~\citep{zhao2026opsd} and SDPO~\citep{hubotter2026sdpo} minimize per-token KL divergence between a privileged teacher $P_T^+$ and the student, producing a gradient of the form:
\begin{equation}
\label{eq:opsd_grad}
\nabla_\theta \mathcal{L}_{\text{OPSD}} = -\sum_{v \in \mathcal{V}} P_T^+(v \mid r^+)\,\nabla_\theta \log P_S(v).
\end{equation}
This vocabulary-wide sum encodes $r^+$ directly into every gradient direction. \citep{yang2026rlsd} showed this produces a harmful deviation $\delta(\theta; r^+)$ with variance $\propto I(Y_t; R^+ \mid X)$ that dominates as training progresses, a pathology termed \emph{information leakage} that is irreducible regardless of implementation. Our results confirm it empirically: OPSD and SDPO fall below the untrained baseline on four of five benchmarks (\S\ref{sec:results}).

RLSD~\citep{yang2026rlsd} resolves leakage by evaluating the teacher signal only at the sampled token $y_t$ under stop-gradient, using the evidence ratio $P_T^{+}(y_t)/P_S(y_t)$ solely to modulate the \emph{magnitude} of the GRPO advantage:
\begin{equation}
\label{eq:rlsd_weight}
\begin{split}
w_t^{\text{RLSD}} &= \exp\!\bigl(\operatorname{sign}(A) \cdot
    \operatorname{sg}(\log P_T^{+}(y_t) - \log P_S(y_t))\bigr), \\
\hat{A}_t^{(i)} &= A^{(i)} \cdot \left[(1-\lambda) + \lambda \cdot
    \operatorname{clip}\!\left(w_t^{\text{RLSD}},\, 1{-}\epsilon_w,\, 1{+}\epsilon_w\right)\right].
\end{split}
\end{equation}
Because $\hat{A}_t$ is $\theta$-constant via \text{sg}, no vocabulary-wide sum
appears in the gradient and the update direction is anchored to the verifier.

\subsection{Limitations of Single-Reference Evidence}
\label{sec:rlsd_limitations}

Despite its safety guarantees, RLSD's ratio $P_T^+(y_t)/P_S(y_t)$ has three signal
quality limitations. \textbf{(1) Fluency confound:} the denominator $P_S(y_t)$
reflects base-rate corpus frequency, not semantic relevance, suppressing the ratio at
common tokens regardless of the numerator. \textbf{(2) Asymmetric negative signal:}
for wrong trajectories, the weight $P_S/P_T^+$ penalizes tokens that $r^+$ would have supported, indirect, with no grounding in what $r^-$ predicts. \textbf{(3) One-sided evidence:} $P_T^+/P_S$ cannot distinguish a filler token (supported equally by both $r^+$ and $r^-$) from a decisive reasoning step ($r^+$ supports it, $r^-$ disfavors it); both receive identical weight if their $P_T^+/P_S$ ratio coincides.

\subsection{Contrastive Evidence Policy Optimization}
\label{sec:cepo}

\paragraph{Contrastive evidence delta.}
We replace $P_T^+(y_t)/P_S(y_t)$ with the \emph{contrastive ratio}
$P_T^+(y_t)/P_T^-(y_t)$, where $r^-$ is the final answer of the lowest-reward
rejected rollout in $\mathcal{G}^-$, available at no additional inference cost.
The student prior $P_S$ cancels entirely, eliminating the fluency confound by
construction. The contrastive evidence delta is:
\begin{equation}
\label{eq:ce_delta}
\Delta_t^{\text{CE}} = \operatorname{sg}\!\left(\log \frac{P_T^+(y_t)}{P_T^-(y_t)}\right).
\end{equation}

\paragraph{Bayesian interpretation.}
Applying Theorem~4 of~\citep{yang2026rlsd} to both teachers and subtracting, $P_S$
cancels and we obtain:
\begin{equation}
\label{eq:diff_belief}
\Delta_t^{\text{CE}} =
    \underbrace{\log \frac{P(r^+ \mid x, y_{\leq t})}{P(r^+ \mid x, y_{<t})}}_{\text{belief update for }r^+}
    -
    \underbrace{\log \frac{P(r^- \mid x, y_{\leq t})}{P(r^- \mid x, y_{<t})}}_{\text{belief update for }r^-}.
\end{equation}
Thus $\Delta_t^{\text{CE}}$ is the \emph{differential belief update}: how much token $y_t$ simultaneously strengthens posterior belief in $r^+$ and weakens it for $r^-$. Decisive steps receive large positive $\Delta_t^{\text{CE}}$; filler tokens receive $\Delta_t^{\text{CE}} \approx 0$.

\paragraph{Token-level advantage and update.}
The contrastive weight and clipped token-level advantage are:
\begin{equation}
\label{eq:ce_weight}
\begin{split}
w_t^{\text{CE}} &= \exp\!\bigl(\operatorname{sign}(A) \cdot \Delta_t^{\text{CE}}\bigr)
    = \left(\frac{P_T^+(y_t)}{P_T^-(y_t)}\right)^{\!\operatorname{sign}(A)}, \\
\hat{A}_t^{(i)} &= A^{(i)} \cdot \bigl[(1{-}\lambda) + \lambda \cdot
    \operatorname{clip}(w_t^{\text{CE}},\, 1{-}\epsilon_w,\, 1{+}\epsilon_w)\bigr].
\end{split}
\end{equation}
where $\lambda$ decays linearly from $\lambda_0$ to 0 over $T_{\mathrm{warm}}$ steps. The policy is updated by maximizing the standard PPO-style clipped surrogate objective~\citep{schulman2017ppo} with $\hat{A}_t^{(i)}$ in place of $A^{(i)}$. When $\mathcal{G}^- = \emptyset$, we set $P_T^- = P_S$, recovering RLSD exactly. CEPO adds one teacher forward pass over RLSD per trajectory, the same marginal overhead as RLSD over GRPO, with no additional sampling cost. Algorithm~\ref{alg:cepo} summarizes the full procedure.

\begin{algorithm}[t]
\caption{Contrastive Evidence Policy Optimization (CEPO)}
\label{alg:cepo}
\begin{algorithmic}[1]
\Require Policy $\pi_\theta$, dataset $\mathcal{S}$, verifier $R$, group size $G$,
         $\lambda$ schedule, clip bounds $\epsilon_w$, $\epsilon$
\For{each training iteration}
    \For{each $(x, r^+)$ in batch}
        \State Sample $\{y^{(i)}\}_{i=1}^G \sim \pi_\theta(\cdot \mid x)$;
               compute $A^{(i)}$ via Eq.~\eqref{eq:grpo_advantage}
        \State $r^- \leftarrow \operatorname{answer}\!\bigl(\arg\min_{j \in \mathcal{G}^-} R(y^{(j)})\bigr)$;
               if $\mathcal{G}^- = \emptyset$ set $P_T^- \leftarrow P_S$
        \For{each trajectory $i$, each position $t$}
            \State $\Delta_t \leftarrow \operatorname{sg}(\log P_T^+(y_t) - \log P_T^-(y_t))$
            \State $\hat{A}_t^{(i)} \leftarrow A^{(i)} \cdot [(1{-}\lambda) +
                   \lambda \cdot \operatorname{clip}(e^{\operatorname{sign}(A^{(i)})\Delta_t},\,
                   1{-}\epsilon_w,\, 1{+}\epsilon_w)]$
        \EndFor
        \State Update $\theta$ via PPO clipped surrogate with $\hat{A}_t^{(i)}$
    \EndFor
\EndFor
\end{algorithmic}
\end{algorithm}

\paragraph{Theoretical guarantees.}
We establish three formal properties of CEPO (proofs in Appendix~\ref{app:proofs}).

\begin{theorem}[CEPO Properties]
\label{thm:cepo}
For $\lambda \in [0,1]$ and $\epsilon_w \in (0,1)$, CEPO satisfies:
\begin{enumerate}[label=(\roman*),nosep,leftmargin=*]
\item \textbf{Direction anchoring.} $\operatorname{sign}(\hat{A}_t) = \operatorname{sign}(A)$ for all $t$, privileged information cannot flip any token's update direction.
\item \textbf{Leakage-free gradient.} $\nabla_\theta \mathcal{L}_{\mathrm{CEPO}}$ contains no vocabulary-wide $r$-conditioned sum; $r^+$ and $r^-$ enter only as stop-gradiented scalars at the sampled token.
\item \textbf{RLSD containment.} Setting $P_T^- = P_S$ recovers RLSD exactly; RLSD is the degenerate case where the wrong-answer teacher carries no information.
\end{enumerate}
\end{theorem}

Beyond safety, we characterize \emph{when} CEPO strictly improves over RLSD.

\begin{proposition}[Discriminative sharpness]
\label{prop:sharpness}
For a correct trajectory: $w_t^{\mathrm{CE}} > w_t^{\mathrm{RLSD}}$ if and only if $P_T^-(y_t) < P_S(y_t)$, precisely when the wrong-answer teacher disfavors this token relative to the student prior. The symmetric condition holds for wrong trajectories. At filler tokens, $P_T^-$ and $P_T^+$ both track $P_S$ closely, so $w_t^{\mathrm{CE}} \approx w_t^{\mathrm{RLSD}} \approx 1$: CEPO introduces no spurious signal where none is warranted.
\end{proposition}

This concentration property is the crux of CEPO's design. RLSD's denominator $P_S(y_t)$ is blind to $r^-$, so it cannot distinguish a decisive reasoning step from a fluent filler token when both happen to have the same $P_T^+/P_S$ ratio. CEPO's denominator $P_T^-$ breaks this tie: a token the wrong answer actively disfavors receives a smaller denominator and strictly higher credit, exactly at positions where the gradient signal is semantically meaningful. The filler-token neutrality is therefore not a limitation but a correctness criterion, amplifying filler gradients would introduce noise, not signal. We validate the sharpness conditions empirically via token-weight analysis in \S\ref{sec:analysis}.\footnote{CEPO is not equivalent to a contrastive KL objective: the gradient of $D_{\mathrm{KL}}(P_T^+\|P_S) - D_{\mathrm{KL}}(P_T^-\|P_S)$ produces a vocabulary-wide sum $-\sum_v [P_T^+(v) - P_T^-(v)]\nabla_\theta \log P_S(v)$, structurally identical to OPSD's leakage flaw (Eq.~\ref{eq:opsd_grad}).}
 
\begin{table}[t]
\centering
\caption{\textbf{Results on five multimodal mathematical reasoning benchmarks.}
  All methods trained 50 steps on Geo3k, $\lambda_0{=}0.5$, $\epsilon_w{=}0.5$. OPSD and SDPO degradation below baseline is consistent with the information leakage identified in \S\ref{sec:background}.}
\label{tab:main}
\small
\setlength{\tabcolsep}{6pt}
\resizebox{\linewidth}{!}{
\begin{tabular}{llcccccr}
\toprule
& & \textbf{DynaMath} & \textbf{LogicVista} & \textbf{MathVis.}$_\mathbf{m}$ 
  & \textbf{MMMU} & \textbf{WeMath} & \textbf{Average Acc.} \\
\midrule
\rowcolor{white}
\multicolumn{8}{l}{\textit{Qwen3-VL-2B-Instruct}} \\
\rowcolor{verylightgray}
& Base  & \res{50.08}{0.7} & \res{32.81}{2.2} & \res{19.41}{2.3} & \res{44.11}{1.7} & \res{52.24}{1.2} & \res{39.73}{0.8} \\
\rowcolor{white}
& +GRPO~\citep{shao2024deepseekmath} & \res{50.36}{0.7} & \res{37.50}{2.3} & \res{21.05}{2.3} & \res{42.33}{1.6} & \res{54.60}{1.2} & \res{41.17}{0.8} \\
\rowcolor{verylightgray}
& +OPSD~\citep{zhao2026opsd} & \res{46.85}{0.7} & \res{28.79}{2.1} & \res{14.14}{2.0} & \res{43.78}{1.7} & \res{41.26}{1.2} & \res{34.96}{0.7} \\
\rowcolor{white}
& +SDPO~\citep{hubotter2026sdpo} & \res{46.65}{0.7} & \res{29.46}{2.2} & \res{15.46}{2.1} & \res{43.00}{1.7} & \res{43.91}{1.2} & \res{35.70}{0.8} \\
\rowcolor{verylightgray}
& +RLSD~\citep{yang2026rlsd}  & \res{50.36}{0.7} & \res{36.38}{2.3} & \res{23.39}{2.3} & \res{39.44}{1.6} & \res{55.26}{1.2} & \res{40.05}{0.8} \\
\rowcolor{white}
& \textbf{+CEPO (Ours)} & \best{51.44}{0.7} & \best{37.72}{2.3} & \best{25.99}{2.5} & \best{45.78}{1.7} & \best{56.21}{1.2} & \best{43.43}{0.8} \\
\midrule
\rowcolor{white}
\multicolumn{8}{l}{\textit{Qwen3-VL-4B-Instruct}} \\
\rowcolor{verylightgray}
& Base  & \res{64.59}{0.7} & \res{54.91}{2.4} & \res{44.41}{2.8} & \res{53.56}{1.7} & \res{74.31}{1.0} & \res{58.36}{0.8} \\
\rowcolor{white}
& +GRPO~\citep{shao2024deepseekmath} & \res{63.97}{0.7} & \res{54.98}{2.4} & \res{42.76}{2.8} & \res{52.34}{1.7} & \res{73.10}{1.1} & \res{57.43}{0.9} \\
\rowcolor{verylightgray}
& +OPSD~\citep{zhao2026opsd} & \res{61.80}{0.7} & \res{55.58}{2.3} & \res{44.41}{2.8} & \res{47.00}{1.7} & \res{72.36}{1.1} & \res{56.23}{0.8} \\
\rowcolor{white}
& +SDPO~\citep{hubotter2026sdpo} & \res{61.58}{0.7} & \res{52.01}{2.4} & \res{43.42}{2.8} & \res{48.11}{1.7} & \res{73.62}{1.1} & \res{55.75}{0.9} \\
\rowcolor{verylightgray}
& +RLSD~\citep{yang2026rlsd} & \res{65.07}{0.7} & \res{56.92}{2.3} & \res{44.08}{2.8} & \res{53.22}{1.7} & \res{73.28}{1.1} & \res{58.51}{0.8} \\
\rowcolor{white}
& \textbf{+CEPO (Ours)} & \best{65.37}{0.7} & \best{61.16}{2.3} & \best{47.37}{2.9} & \best{54.11}{1.7} & \best{74.77}{1.0} & \best{60.56}{0.9} \\
\bottomrule
\end{tabular}
}
\end{table}

\section{Experiments}\label{sec:experiments}

\begin{wraptable}{r}{0.30\linewidth}
\vspace{-1.8em}
\centering
\caption{\textbf{Wall-clock training time} for 50 steps on Geo3k.}
\label{tab:wallclock}
\small
\rowcolors{2}{verylightgray}{white}
\begin{tabularx}{\linewidth}{Xc}
\toprule
\textbf{Method} & \textbf{Time} \\
\midrule
GRPO  & 5h 58m \\
SDPO  & 6h 14m \\
RLSD  & 6h 15m \\
CEPO  & 6h 34m \\
\bottomrule
\end{tabularx}
\vspace{-1.0em}
\end{wraptable}
\paragraph{Models and training.}
We train Qwen3-VL-2B-Instruct and Qwen3-VL-4B-Instruct~\citep{qwen3vl}
using the EasyR1~\citep{easyr1} framework with FSDP~\citep{fsdp} and
vLLM~\citep{vllm}-accelerated inference. All models are fine-tuned with LoRA (rank 16)
for 50 steps on Geo3k~\citep{geo3k}, a geometry question-answering
dataset of 3,000 training problems with verifiable numeric answers.
We use AdamW~\citep{adamw} with lr $10^{-6}$ (CEPO $5\!\times\!10^{-6}$), batch size 32, rollout group
size $G = 8$, and maximum sequence length 2,048 tokens.
For all CEPO runs, $\lambda_0 = 0.5$ with linear decay to 0 over
$T_{\mathrm{warm}} = 25$ steps and $\epsilon_w = 0.5$ unless otherwise stated.
The negative reference $r^-$ is the final answer extracted from the lowest-reward rejected rollout in the current group. The teacher is the same as the actor. All experiments run on NVIDIA RTX6000 Pro Blackwell 100GBs GPUs. Table~\ref{tab:wallclock} reports wall-clock training times; CEPO's two
teacher forward passes add 36 minutes over GRPO, comparable to
that of RLSD/SDPO over GRPO.

\paragraph{Baselines.}
We compare against four baselines under identical training budgets: \textbf{GRPO}~\citep{shao2024deepseekmath}, the sequence-level RL baseline; \textbf{OPSD}~\citep{zhao2026opsd}, which minimizes per-token KL divergence to a correct-answer teacher; \textbf{SDPO}~\citep{hubotter2026sdpo}, which extends OPSD with Jensen-Shannon divergence and EMA teacher stabilization; and \textbf{RLSD}~\citep{yang2026rlsd}, the direct predecessor of CEPO. All baselines use the same LoRA rank, group size, and training steps as CEPO. Other training hyperparameters are detailed in Appendix~\ref{app:hyperparams}.

\paragraph{Evaluation.}
We report accuracy on five held-out multimodal mathematical reasoning benchmarks: DynaMath~\citep{dynamath},LogicVista~\citep{logicvista}, MathVision$_{\mathrm{mini}}$~\citep{mathvision}, MMMU~\citep{mmmu}, and WeMath~\citep{wemath}. All models are evaluated using lmms-eval~\citep{lmms-eval} with sampling (temperature 1.0, top-$p$ 1.0, top-$k$ 40, presence penalty 2.0, maximum 32,000 tokens). 

\section{Results}\label{sec:results}

Table~\ref{tab:main} reports results on both model scales. On Qwen3-VL-2B, CEPO achieves 43.43\% average accuracy, compared to 41.17\% for GRPO (+2.26pp), 34.96\% for OPSD, and 35.70\% for SDPO. On Qwen3-VL-4B, CEPO achieves 60.56\%, versus 57.43\% for GRPO (+3.13pp) and 56.23\% for OPSD. Gains are most pronounced on LogicVista (+6.18pp over GRPO on 4B) and MathVision$_\mathrm{mini}$ (+4.94pp over GRPO on 2B), benchmarks that reward fine-grained multi-step reasoning over short, pattern-matchable answers. MMMU, which is primarily a multiple-choice knowledge retrieval benchmark with limited reasoning chains, shows the smallest gain (+1.67pp on 2B), consistent with the expectation that CEPO's contrastive signal provides less leverage when reasoning traces are short.

\paragraph{OPSD and SDPO degradation.}
A notable finding is that both OPSD and SDPO fall \emph{below} the untrained base model on 2B (34.96\% and 35.70\% vs.\ 39.73\%). This is consistent with the information leakage analysis in \S\ref{sec:background}: as training progresses, the vocabulary-wide $r$-conditioned gradient deviation $\delta(\theta; r^+)$ dominates the benign signal, driving the model to encode spurious $x \to r^+$ correlations that degrade generalization. The same pattern appears at 4B (56.23\% for OPSD vs.\ 58.36\% base), confirming that the leakage pathology is not an artifact of model scale. CEPO avoids this entirely: its gradient contains no vocabulary-wide $r$-conditioned term by construction (Theorem~\ref{thm:cepo}(ii)).

\subsection{Ablations}
\label{sec:ablations}

\begin{wraptable}{r}{0.48\linewidth}
\vspace{-1.5em}
\centering
\caption{\textbf{Teacher source ablation.} $\Delta$ is relative to GRPO.}
\vspace{0.5em}
\label{tab:teacher}
\small
\setlength{\tabcolsep}{5pt}
\rowcolors{2}{verylightgray}{white}
\begin{tabular}{lcc}
\toprule
\textbf{Teacher Source} & \textbf{Avg. Acc} & $\boldsymbol{\Delta}$\textbf{GRPO} \\
\midrule
\textit{GRPO baseline} & $41.17_{\pm0.8}$ & $0.00$ \\
Reference policy & $42.18_{\pm0.8}$ & $+1.01$ \\
Sync every 25 steps & $42.74_{\pm0.8}$ & $+1.57$ \\
Actor policy & $\mathbf{43.43}_{\pm0.8}$ & $\mathbf{+2.26}$ \\
\bottomrule
\end{tabular}
\vspace{-1.0em}
\end{wraptable}

\paragraph{Teacher source (Table~\ref{tab:teacher}).}
We compare three teacher sources: a fixed reference policy, a periodically synced teacher, and the actor policy itself. The actor-policy teacher performs best, reaching 43.43\%, a +2.26pp improvement over GRPO. This indicates that, in our setting, the most useful teacher is the one aligned with the current on-policy rollout distribution, even if its token distribution remains close to the student. Crucially, sharing weights with the actor requires no separate parameter copy, reducing memory overhead. The fixed reference policy improves over GRPO but reaches only 42.18\%, suggesting that a frozen teacher provides a useful but increasingly stale contrastive signal as the policy changes. Synchronizing the teacher with the actor every 25 steps improves performance to 42.74\%, narrowing the gap to the actor-policy teacher by keeping the teacher fresher while still partially decoupling it from the student. Overall, these results suggest that teacher freshness and on-policy alignment are more important than maintaining a large teacher-student distribution gap for CEPO.

\begin{table}[t]
\centering
\caption{\textbf{Feedback source ablation.}
Average accuracy on Qwen3-VL-2B after 50 Geo3k training steps; $\Delta$ is relative to GRPO.}
\label{tab:feedback}
\small
\setlength{\tabcolsep}{6pt}
\rowcolors{2}{verylightgray}{white}
\begin{tabular}{llcc}
\toprule
\textbf{Positive} $r^+$ & \textbf{Negative} $r^-$ 
  & \textbf{Avg. Acc} & $\boldsymbol{\Delta}$\textbf{GRPO} \\
\midrule
\multicolumn{2}{l}{\textit{GRPO baseline, no teacher}} & $41.17_{\pm0.8}$ & $0.00$ \\
Peer rollout (prefix) & Peer rollout (prefix)       & $40.47_{\pm0.8}$ & $-0.70$ \\
Peer rollout (suffix) & Peer rollout (suffix)       & $40.60_{\pm0.8}$ & $-0.57$ \\
Peer rollout (full)   & Peer rollout (full)         & $41.99_{\pm0.8}$ & $+0.82$ \\
Ground truth answer   & Peer rollout (full)         & $42.74_{\pm0.8}$ & $+1.57$ \\
Ground truth answer   & Peer rollout (answer only)  & $\mathbf{43.43}_{\pm0.8}$ & $\mathbf{+2.26}$ \\
\bottomrule
\end{tabular}
\end{table}

\paragraph{Feedback source (Table~\ref{tab:feedback}).}
We ablate the construction of $r^+$ and $r^-$ across five configurations. The main CEPO setting, ground truth final
answer as $r^+$ and peer answer only as $r^-$, performs best at 43.43\%,
improving over GRPO by +2.26pp. Using the full peer rollout as the
negative reference also improves performance, reaching 42.74\%, while
full peer rollout conditioning on both sides reaches 41.99\%. Partial peer context performs worse. Prefix only and suffix only
conditioning reach 40.47\% and 40.60\%, both below GRPO, suggesting that
truncated reasoning traces provide a noisy contrastive signal. Overall,
the strongest ablation result comes from using the verified final answer
as the positive reference and a compact rejected answer as the negative
reference.

\begin{figure}[t]
    \centering
    \includegraphics[width=\linewidth]{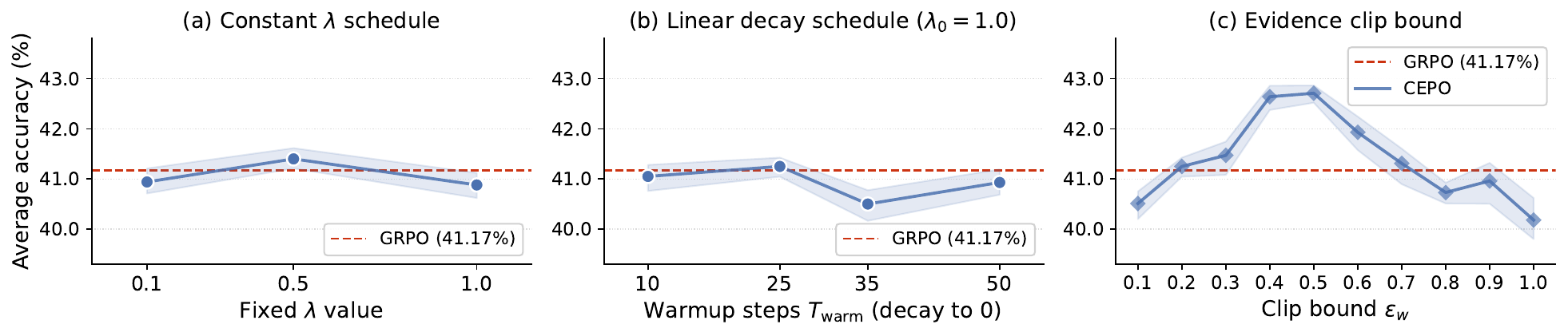}

    \caption{\textbf{Hyperparameter sensitivity} averaged across 5 reasoning benchmarks).
    \textbf{(a)} Constant $\lambda$ schedule: $\lambda = 0.5$ peaks at
    $41.40\%$, outperforming GRPO ($41.17\%$); sustained high-$\lambda$ training
    ($\lambda = 1.0$) introduces noise that offsets the credit-assignment benefit.
    \textbf{(b)} Linear-decay schedule from $\lambda_0 = 1.0$: a $25$-step warmup
    matches the constant-$\lambda$ peak ($41.25\%$).
    \textbf{(c)} Evidence clip bound $\varepsilon_w$: performance peaks in
    $[0.4, 0.5]$ at $42.7\%$ ($+1.5$pp over GRPO) and degrades at both extremes,
    small $\varepsilon_w$ collapses CEPO toward GRPO, large $\varepsilon_w$
    destabilizes the modulated advantage.}
    
    \label{fig:ablation_hyperparams}
\end{figure}
\paragraph{Hyperparameter sensitivity (Figure~\ref{fig:ablation_hyperparams}).}
\textit{Evidence clip bound $\epsilon_w$.}
Performance peaks at $\epsilon_w \in [0.4, 0.5]$ and degrades toward
both extremes. At $\epsilon_w = 0.1$, the clip is too tight and the
method effectively reduces to GRPO. At $\epsilon_w \geq 0.8$, unconstrained
weights introduce variance that destabilizes advantage estimation.
We recommend $\epsilon_w = 0.5$ as the default. \textit{$\lambda$ schedule.}
A constant $\lambda = 0.5$ and a 25-step linear decay both outperform
GRPO, while $\lambda = 1.0$ (constant maximum) performs worse despite the
highest integrated CEPO pressure (50 units vs.\ 25 for $\lambda = 0.5$).
A 10-step fast decay achieves comparable performance to the 25-step
schedule, suggesting that the benefit of contrastive credit assignment is
front-loaded: the first 10--25 steps drive the bulk of the improvement.
Extending the schedule beyond 25 steps introduces noise that offsets the
signal.

\subsection{Analysis}
\label{sec:analysis}

\begin{wrapfigure}{r}{0.48\linewidth}
    \vspace{-1.2em}
    \centering
    \includegraphics[width=\linewidth]{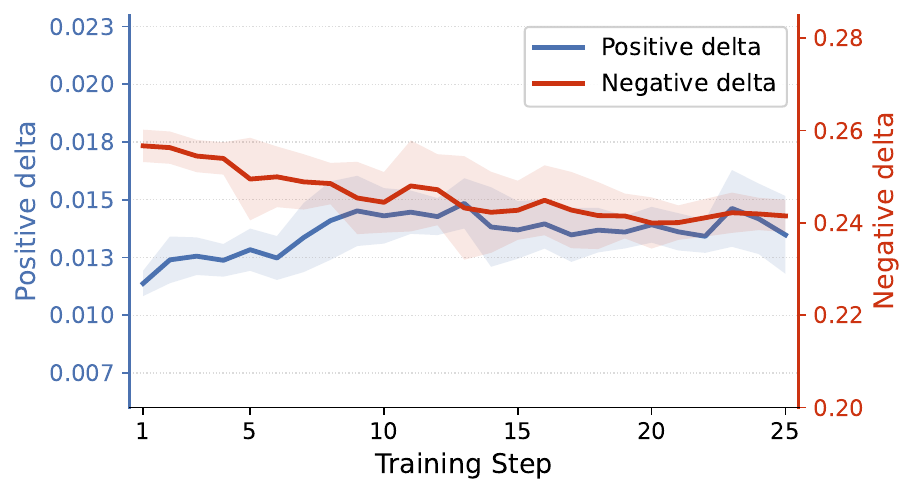}
    \caption{\textbf{Contrastive delta fractions during CEPO training.}
    We track the fraction of tokens assigned positive versus negative
    contrastive evidence. Positive-delta mass increases early, while
    negative-delta mass decreases.}
    \label{fig:delta_fraction}
    \vspace{-1.0em}
\end{wrapfigure}

\paragraph{Contrastive delta fractions.}
Figure~\ref{fig:delta_fraction} tracks the sign structure of CEPO's raw
contrastive delta $\Delta_t^{\mathrm{CE}}=\log P_T^+(y_t)-\log P_T^-(y_t)$
before clipping or reweighting. A positive delta means the token
is more supported by the correct-answer teacher, so CEPO amplifies credit
on positive-advantage rollouts; a negative delta means the token is more
supported by the wrong-answer teacher, so CEPO assigns stronger blame on
negative-advantage rollouts. The positive-delta fraction rises during early
updates, indicating CEPO increasingly identifies tokens supporting
correct reasoning, while the negative-delta fraction declines, suggesting
the model produces fewer tokens compatible with the rejected-answer teacher.
This confirms CEPO's intended behavior: training shifts evidence
toward correct-answer support rather than uniformly increasing weights.

\begin{figure}
    \centering
    \includegraphics[width=\linewidth]{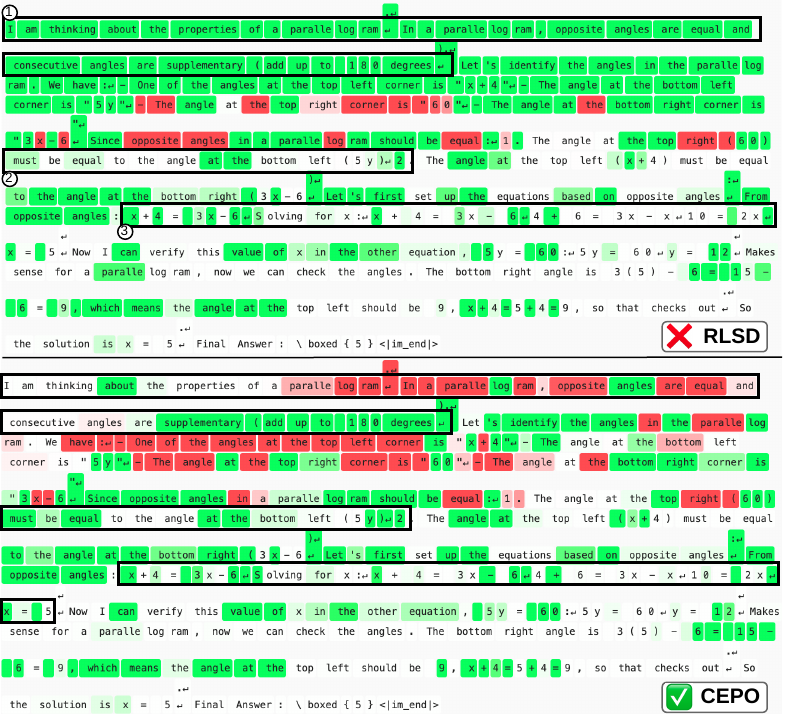}
    \caption{\textbf{Token-level credit assignment on a parallelogram problem}. Green/red/white denote high, low, and neutral token weights. Numbered regions illustrate three claims: \ding{172} RLSD over-credits fluent setup prose, while CEPO suppresses it; \ding{173} CEPO localizes blame to the misapplied angle-equality inference instead of diffusing penalties; \ding{174} CEPO sharpens credit on the decisive algebraic derivation ($x{+}4=3x{-}6$, isolation steps, final answer). The lower CEPO clip rate (49.5\% vs.\ 71.3\%) indicates a wider effective dynamic range, consistent with Proposition~\ref{prop:sharpness}.}
    \label{fig:heatmap}
\end{figure}

\paragraph{Token weight heatmap.}
Figure~\ref{fig:heatmap} compares token-level credit assignment between RLSD and CEPO on the same geometry trajectory. In the top (RLSD) panel, credit is spread broadly across fluent setup prose and connective tokens, with no strong concentration on decisive steps. In the bottom (CEPO) panel, the contrastive signal sharpens credit onto the critical algebraic derivation and final answer tokens, while suppressing filler to near-unity weight. The lower CEPO clip rate (49.5\% vs.\ 71.3\%) confirms that CEPO operates over a wider effective dynamic range than RLSD. This qualitative pattern is consistent with Proposition~\ref{prop:sharpness} and provides interpretable evidence that the contrastive denominator $P_T^-$ successfully distinguishes decisive reasoning steps from fluent but uninformative tokens.

\section{Conclusion}
\label{sec:conclusion}

We presented CEPO, a token-level credit assignment method for RLVR that replaces RLSD's single-reference evidence ratio with a contrastive ratio between correct- and wrong-answer teachers drawn from rejected rollouts in the training batch. We proved this preserves all structural safety guarantees of RLSD (direction anchoring and leakage-free gradients) while strictly sharpening credit at decisive tokens and leaving filler unchanged (Theorem~\ref{thm:cepo}, Proposition~\ref{prop:sharpness}). CEPO outperforms GRPO, OPSD, and SDPO across five multimodal mathematical reasoning benchmarks at 2B and 4B scale; the collapse of OPSD and SDPO below the untrained baseline confirms that structural safety is a practical prerequisite, not a theoretical nicety. These results are validated on Qwen3-VL trained on Geo3k, and extending CEPO to larger models, text-only reasoning, and code generation is a natural next step. We hope CEPO offers a principled and practical building block for the next generation of credit-aware RLVR training.

\bibliographystyle{plain}
\bibliography{references}


\appendix
\clearpage
{\huge \textbf{Appendix}}
\section{Proofs}
\label{app:proofs}

\subsection{Proof of Theorem~\ref{thm:cepo}}

\paragraph{(i) Direction anchoring.}
Since $\exp(\cdot) > 0$, we have $w_t^{\mathrm{CE}} > 0$ unconditionally.
Because $\epsilon_w \in (0,1)$, both clip bounds $1 \pm \epsilon_w$ are positive,
so $\operatorname{clip}(w_t^{\mathrm{CE}}, 1{-}\epsilon_w, 1{+}\epsilon_w) > 0$.
For any $\lambda \in [0,1]$:
\[
(1-\lambda) + \lambda \cdot \operatorname{clip}(w_t^{\mathrm{CE}}, 1{-}\epsilon_w, 1{+}\epsilon_w) > 0,
\]
since it is a convex combination of $1$ and a positive quantity.
Therefore $\hat{A}_t = A \cdot [\text{positive}]$, giving
$\operatorname{sign}(\hat{A}_t) = \operatorname{sign}(A)$ unconditionally. \hfill$\square$

\paragraph{(ii) Leakage-free gradient.}
The stop-gradient on $\Delta_t^{\mathrm{CE}}$ renders $\hat{A}_t^{(i)}$
$\theta$-constant within each update step. Therefore:
\[
\nabla_\theta \mathcal{L}_{\mathrm{CEPO}}
= \mathbb{E}\!\left[\frac{1}{G}\sum_i \frac{1}{|y^{(i)}|}\sum_t
  \hat{A}_t^{(i)} \cdot \nabla_\theta \log \pi_\theta(y_t^{(i)} \mid x, y_{<t}^{(i)})\right],
\]
where $\hat{A}_t^{(i)}$ is constant. The gradient acts only at the sampled token
$y_t^{(i)}$; no vocabulary-wide sum $\sum_{v \in \mathcal{V}}$ appears, and $r^+, r^-$
enter only through the $\theta$-constant scalar $\hat{A}_t^{(i)}$. \hfill$\square$

\paragraph{(iii) RLSD containment.}
When $P_T^-(y_t) = P_S(y_t)$ for all $t$:
\[
\Delta_t^{\mathrm{CE}}
= \log P_T^+(y_t) - \log P_T^-(y_t)
= \log P_T^+(y_t) - \log P_S(y_t)
= \Delta_t^{\mathrm{RLSD}}.
\]
Hence $w_t^{\mathrm{CE}} = w_t^{\mathrm{RLSD}}$ and
$\hat{A}_t^{\mathrm{CEPO}} = \hat{A}_t^{\mathrm{RLSD}}$ for all $t$. \hfill$\square$

\subsection{Proof of Proposition~\ref{prop:sharpness}}

From Eqs.~\eqref{eq:ce_weight} and~\eqref{eq:rlsd_weight}:
\[
w_t^{\mathrm{CE}}\big|_{A>0} = \frac{P_T^+(y_t)}{P_T^-(y_t)}, \qquad
w_t^{\mathrm{RLSD}}\big|_{A>0} = \frac{P_T^+(y_t)}{P_S(y_t)}.
\]

\begin{table}[t]
\centering
\caption{\textbf{Shared training hyperparameters.} Identical across all five methods
in our experimental setup.}
\label{tab:hyperparams_shared}
\small
\rowcolors{2}{verylightgray}{white}
\begin{tabular}{ll}
\toprule
\textbf{Hyperparameter} & \textbf{Value} \\
\midrule
Base models          & Qwen3-VL-2B-Instruct, Qwen3-VL-4B-Instruct \\
Training dataset     & Geo3k (3{,}000 geometry problems) \\
Training steps       & 50 \\
Optimizer            & AdamW \\
Learning rate        & $1 \times 10^{-6}$ ($5 \times 10^{-6}$ for CEPO)\\
LR schedule          & Cosine decay, 5-step linear warmup \\
Batch size (prompts) & 32 \\
Rollout group size $G$ & 8 \\
Rollout temperature  & 1.0 \\
LoRA rank / $\alpha$ & 16 / 32 \\
LoRA dropout         & 0.0 \\
Max sequence length  & 2{,}048 tokens \\
PPO clip bound high $\epsilon_{high}$ & 0.28 \\
PPO clip bound low $\epsilon_{low}$ & 0.20 \\
KL penalty           & None \\
Entropy regularization & None \\
Reward model         & Rule-based verifier (numeric match) \\
Framework            & EasyR1 + FSDP + vLLM \\
\bottomrule
\end{tabular}
\end{table}

\paragraph{Case 1 ($A > 0$).}
$w_t^{\mathrm{CE}} > w_t^{\mathrm{RLSD}}$ iff
$P_T^+(y_t)/P_T^-(y_t) > P_T^+(y_t)/P_S(y_t)$.
Since $P_T^+(y_t) > 0$, this reduces to $1/P_T^-(y_t) > 1/P_S(y_t)$,
i.e.\ $P_S(y_t) > P_T^-(y_t)$. Under the joint condition
$P_T^+(y_t) \geq P_S(y_t)$, RLSD already assigns above-baseline credit
($w_t^{\mathrm{RLSD}} \geq 1$) and CEPO strictly amplifies it. \hfill$\square$

\paragraph{Case 2 ($A < 0$).}
The weights are $w_t^{\mathrm{CE}} = P_T^-(y_t)/P_T^+(y_t)$ and
$w_t^{\mathrm{RLSD}} = P_S(y_t)/P_T^+(y_t)$.
By the same argument, $w_t^{\mathrm{CE}} > w_t^{\mathrm{RLSD}}$
iff $P_T^-(y_t) > P_S(y_t)$. Under the joint condition
$P_T^+(y_t) \leq P_S(y_t)$, CEPO assigns strictly stronger blame
to a token RLSD already penalizes. \hfill$\square$

\paragraph{Case 3 (filler).}
When $P_T^+(y_t) \approx P_T^-(y_t) \approx P_S(y_t)$, all ratios
are near 1, so $\Delta_t^{\mathrm{CE}} \approx 0$ and
$w_t^{\mathrm{CE}} \approx w_t^{\mathrm{RLSD}} \approx 1$.
Neither method discriminates at informationally neutral positions. \hfill$\square$

\section{Baseline Hyperparameter Details}
\label{app:hyperparams}

All methods in our experiments share a common training infrastructure based on
Qwen3-VL-\{2B,4B\}-Instruct fine-tuned with LoRA via the EasyR1~\citep{easyr1} framework with
FSDP and vLLM-accelerated rollout generation.
Table~\ref{tab:hyperparams_shared} reports the shared infrastructure hyperparameters
and Table~\ref{tab:hyperparams_specific} reports the method-specific hyperparameters.

\begin{table}[t]
\centering
\caption{\textbf{Method-specific hyperparameters} used in our reimplementation.
  ``---'' denotes parameters not applicable to a given method.}
\label{tab:hyperparams_specific}
\small
\setlength{\tabcolsep}{4pt}
\resizebox{\linewidth}{!}{
\rowcolors{2}{verylightgray}{white}
\begin{tabular}{lccccc}
\toprule
\textbf{Hyperparameter}
  & \textbf{GRPO} & \textbf{OPSD} & \textbf{SDPO} & \textbf{RLSD} & \textbf{CEPO} \\
\midrule
Distillation objective
  & --- & $D_{\mathrm{KL}}(P_T^+\!\parallel\! P_S)$ & $D_{\mathrm{JS}}(P_T^+\!\parallel\! P_S)$ & --- & --- \\
Distillation weight $\alpha$
  & --- & 1.0 & 1.0 & --- & --- \\
EMA teacher decay $\beta$
  & --- & --- & 0.999 & --- & --- \\
EMA update frequency
  & --- & --- & every step & --- & --- \\
Evidence weight $\lambda_0$
  & --- & --- & --- & 0.5 & 0.5 \\
$\lambda$ decay schedule
  & --- & --- & --- & linear $\to 0$ & linear $\to 0$ \\
Warmup steps $T_{\mathrm{warm}}$
  & --- & --- & --- & 25 & 25 \\
Evidence clip $\epsilon_w$
  & --- & --- & --- & 0.5 & 0.5 \\
Positive reference $r^+$
  & --- & ground truth answer & ground truth answer & ground truth answer & ground truth answer \\
Negative reference $r^-$
  & --- & --- & --- & --- & rejected rollout (answer only) \\
Teacher source
  & --- & actor & EMA of actor & actor & actor \\
Teacher sync frequency
  & --- & every step & EMA & every 50 steps & every step \\
Stop-gradient on teacher
  & --- & Yes & Yes & Yes & Yes \\
\bottomrule
\end{tabular}
}
\end{table}

\paragraph{Training prompts.}
The prompt used for training the student is as follows:
\begin{lstlisting}
{{ problem }} Solve the problem step by step, keeping reasoning brief. 
Put ONLY the final answer inside \boxed{}.
\end{lstlisting}

The prompt for computing the log probabilities through the teacher is as follows: 
\begin{lstlisting}
{{ problem }} Solve the problem step by step, keeping reasoning brief. 
Put ONLY the final answer inside \boxed{}.
Here is a sample answer: {{ ground_truth_answer }}
{{ student_answer }}
\end{lstlisting}




\end{document}